%% file: iswc14-all.tex
\documentclass[runningheads,a4paper]{llncs}

\usepackage{amssymb}
\usepackage{graphicx}
\usepackage{epstopdf}
\usepackage{epsfig,graphicx}
\usepackage{color}
\usepackage{url}
\usepackage{tabularx}
\usepackage{array}
\usepackage{multirow}
\usepackage{amsmath}
\usepackage{algorithm,algorithmic}
\usepackage{hhline}
\usepackage{booktabs}

\urldef{\mailsa}\path|{ttran, tunguyen}@L3S.de|

\begin{document}

\linespread{0.9}
\setlength{\parskip}{0cm}
\setlength{\parindent}{1em}

\mainmatter  

\title{Hedera: Scalable Indexing and Exploring Entities in Wikipedia
  Revision History}

\titlerunning{Hedera: Processing Tools for Wikipedia Revision}

%
%
%


\author{Tuan Tran \and Tu Ngoc Nguyen}

\authorrunning{Tuan Tran and Tu Ngoc Nguyen}

\institute{L3S Research Center / Leibniz Universit\"{a}t Hannover, Germany\\\email{\{ttran, tunguyen\}@L3S.de}}

\maketitle
\vspace{-0.2cm}
\begin{abstract}
Much of work in semantic web relying on Wikipedia as the main source of
knowledge often work on static snapshots of
the dataset. The full history of Wikipedia revisions, while contains
much more useful information, is still difficult to access due to its
exceptional volume. To enable further research on this collection, we
developed a tool, named \emph{Hedera}, that efficiently extracts semantic information from
Wikipedia revision history datasets.
Hedera exploits  Map-Reduce paradigm to achieve rapid extraction, it is able
to handle one entire Wikipedia articles' revision history within
a day in a medium-scale cluster, and supports flexible
data structures for various kinds of semantic web study.
\end{abstract}
\vspace{-0.2cm}
\section{Introduction}
\vspace{-0.2cm}
For over decades, Wikipedia has become a backbone of sematic Web research, with the proliferation of high-quality big knowledge bases (KBs) such as DBpedia \cite{auer2007dbpedia}, where information is derived from various Wikipedia public collections. Existing approaches often rely on one offline snapshot of datasets, they treat knowledge as static and ignore the temporal evolution of  information in Wikipedia.
When for instance a fact changes (e.g. death of a celebrity) or entities themselves evolve, they can only be reflected in the next version of the knowledge bases (typically extracted fresh from a newer Wikipedia dump). This undesirable quality of KBs make them unable to capture temporally dynamical relationship latent among revisions of the encyclopedia (e.g., \textit{participate} together in complex events), which are difficult to detect in one single Wikipedia snapshot. Furthermore, applications relying on obsolete facts might fail to reason under new contexts (e.g. question answering systems for recent real-world incidents), because they were not captured in the KBs. In order to complement these temporal aspects, the whole Wikipedia revision history should be well-exploited. However, such longitudinal analytics over ernoumous size of Wikipedia require huge computation. In this work, we develop \emph{Hedera}, a large-scale framework that supports processing, indexing and visualising Wikipedia revision history. Hedera is an end-to-end system that works directly with the raw dataset, processes them to streaming data, and incrementally indexes and visualizes the information of entities registered in the KBs in a dynamic fashion. In contrast to existing work that handle the dataset in centralized settings~\cite{ferschke2011wikipedia}, Hedera employs the Map-Reduce paradigm to achieve the scalable performance, which is able to transfer raw data of  2.5 year revision history of 1 million entities into full-text index within a few hours in an 8-node cluster. We open-sourced Hedera to facilitate further research~\footnote{Project documentation and code can be found at: \url{https://github.com/antoine-tran/Hedera}}.
\vspace{-0.2cm}
\section{Extracting and Indexing Entities}
\subsection{Preprocessing Dataset}
\begin{figure}[ht]
    \begin{center}
        \includegraphics[trim=0cm 8cm 0cm 0.5cm,clip=true,width=0.8\textwidth]{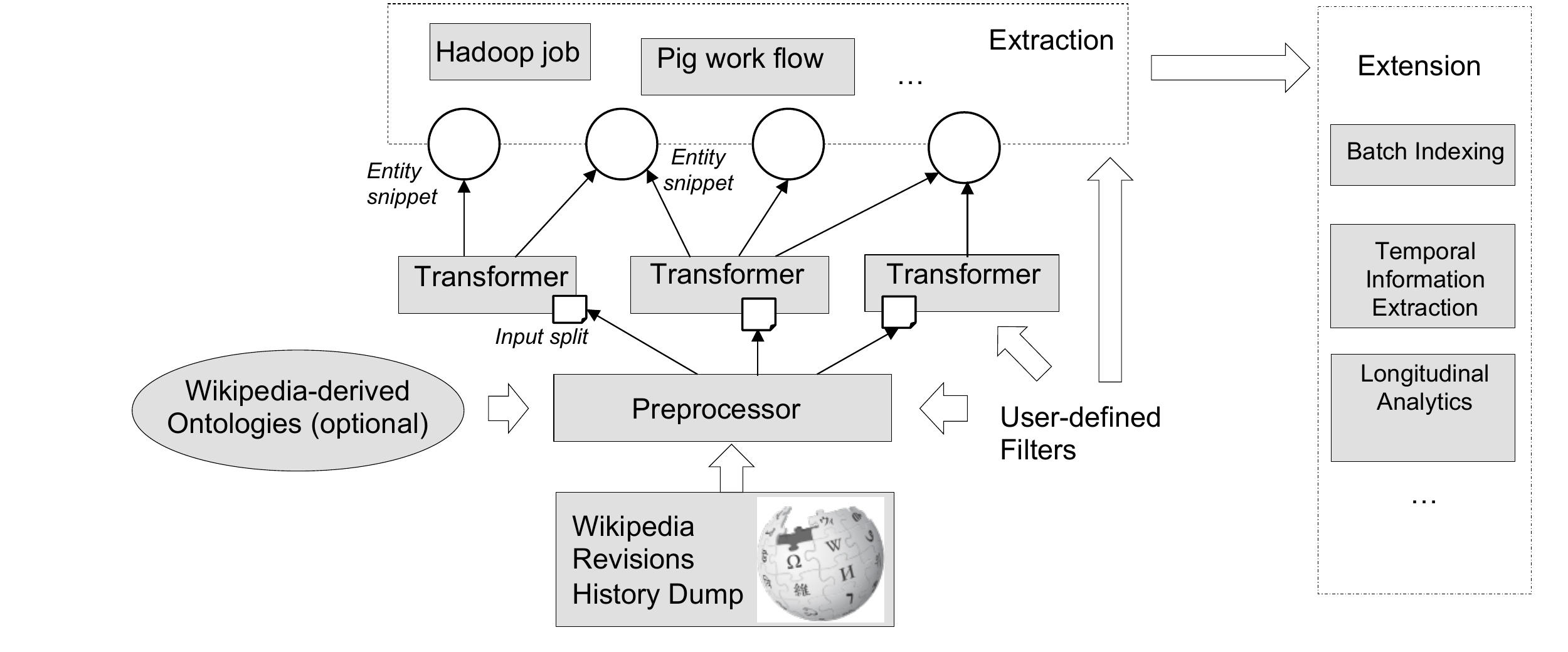}
      \caption{Hedera Framework Architecture}
    \label{fig:architecture}
    \end{center}
\end{figure}

Here we describe the Hedera architecture and workflow. As shown in Figure~\ref{fig:architecture}, the core data input of Hedera is a Wikipedia Revision history dump \footnote{\url{http://dumps.wikimedia.org}}. Hedera currently works with the raw XML dumps, it supports accessing and extracting information directly from compressed files. Hedera makes use heavily the Hadoop framework. The preprocessor is responsible for re-partitioning the raw files into independent units (a.k.a \emph{InputSplit} in Hadoop) depending on users' need. There are two levels of partitioning: Entity-wise and Document-wise. Entity-wise partitioning guarantees that revisions belonging to the same entity are sent to one computing node, while document-wise sends content of revisions arbitrarily to any node, and keeps track in each revision the reference to its preceding ones for future usage in the Map-Reduce level. The preprocessor accepts user-defined low-level filters (for instance, only partition articles, or revisions within 2011 and 2012), as well as list of entity identifiers from a knowledge base to limit to. If filtered by the knowledge base, users must provide methods to verify one revision against the map of entities (for instance, using Wikipedia-derived URL of entities). The results are Hadoop file splits, in the XML or JSON formats.

\vspace{-0.5cm}

\vspace{-1.4cm}
\subsection{Extracting Information}
Before extracted in the Map-Reduce phase (Extraction component in Figure \ref{fig:architecture}), file splits outputed from the preprocessor are streamed into a \textit{Transformer}. The main goal of the transformer is to consume the files and emits (\textit{key},\textit{value}) pairs suitable for inputting into one Map function. Hedera provides several classes of transformer, each of which implements one operator specified in the extraction layer. Pushing down these operators into transformers reduces significantly the volume of text sent around the network. 
The extraction layer enables users to write extraction logic in high-level programming languages such as Java or Pig \footnote{\url{http://pig.apache.org}}, which can be used in other applications. The extraction layer also accepts user-defined filters, allowing user to extract and index different portions of the same partitions at different time. For instance, the user can choose to first filter and partition Wikipedia articles published in 2012; and later she can sample, from one partition, the revisions about people published in May 2012. This flexibility facilitates rapid development of  research-style prototypes in Wikipedia revision dataset, which is one of our major contributions.
\vspace{-0.2cm}
\section{Indexing and Exploring Entity-based Evolutions in Wikipedia}
\label{sec:demo}
In this section, we illustrate the use of Hedera in one application - incremental indexing and visualizing Wikipedia revision history. Indexing large-scale longitudinal data collections i.e., the Wikipedia history is not a straightforward problem. Challenges in finding a scalable data structure and distributed storage that can most exploit data along the time dimension are still not fully addressed. In Hedera, we present a distributed approach in which the collection is processed and thereafter the indexing is parallelized using the Map-Reduce paradigm. This approach (that is based on the document-based data structure of ElasticSearch) can be considered as a baseline for further optimizations. The index's schema is loosely structured, which allows flexible update and incremental indexing of new revisions (that is of necessity for the evolving Wikipedia history collection). Our preliminary evaluation showed that this approach outperformed the well-known centralized indexing method provided by~\cite{ferschke2011wikipedia}. The time processing (indexing) gap is exponentially magnified along with the increase of data volume. In addition, we also evaluated the querying time (and experienced the similar result) of the system. We describe how the temporal index facilitate large-scale analytics on the semantic-ness of Wikipedia with some case studies. The detail of the experiment is described below.

\begin{figure}[ht]
    \begin{center}
      \includegraphics[width=1\columnwidth]{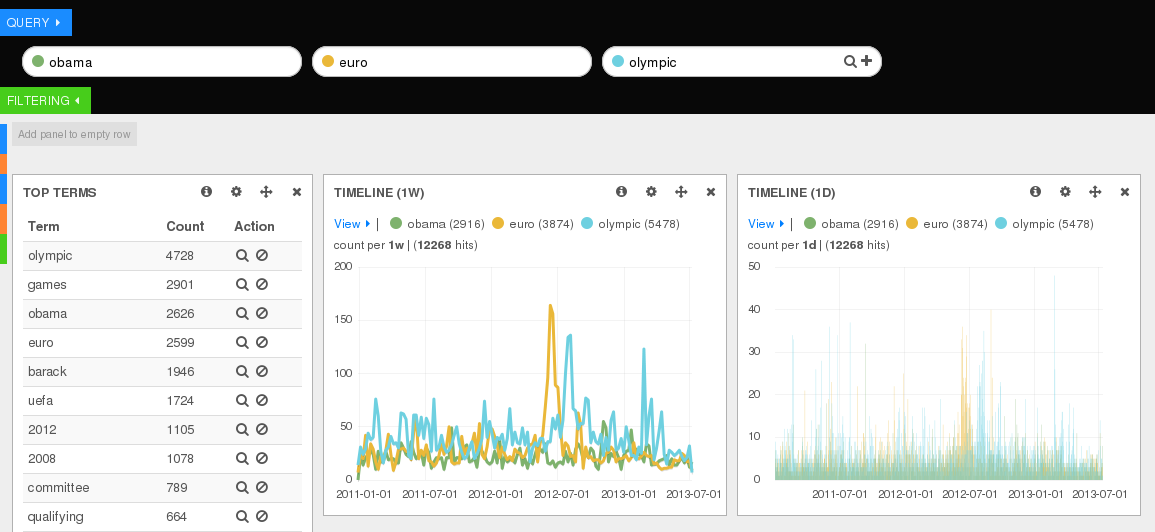}
      \caption{Exploring Entity Structure Dynamics Over Time}
    \label{fig:demo1}
    \end{center}
\end{figure}

We extract 933,837 entities registered in DBpedia, each of which correspond to one Wikipedia article. The time interval spans from 1 Jan 2011 to 13 July 2013, containing 26,067,419 revisions, amounting for 601 GBytes of text in uncompressed format. The data is processed and re-partitioned using Hedera before being passed out and indexed into ElasticSearch \footnote{\url{http://www.elasticsearch.org}} (a distributed real-time indexing framework that supports data at large scale) using Map-Reduce. 
Figure~\ref{fig:demo1} illustrates one toy example of analysing the temporal dynamics of entities in Wikipedia. Here we aggregate the results for three distinct entity queries, i.e., \textit{obama}, \textit{euro} and \textit{olympic} on the temporal \textit{anchor-text} (a visible text on a hyperlink between two Wikipedia revision) index.
The left-most table shows the top terms appear in the returned results, whereas the two timeline graphs illustrate the dynamic evolvement of the entities over the studied time period (with 1-week and 1-day granuality, from left to right respectively). As easily observed, the three entities peak at the time where a related event happens (\textsf{Euro 2012} for \textit{euro}, \textsf{US Presidential Election} for \textit{obama} and the Summer and Winter Olympics for \textit{olympic}). This further shows the value of temporal anchor text in mining the Wikipedia entity dynamics.  We analogously experimented on the  Wikipedia \textit{full-text} index. Here we brought up a case study of the entity co-occurrance (or temporal relationship) (i.e., between \textsf{Usain Bolt} and \textsf{Mo Farah}), where the two co-peak in the time of Summer Olympics 2012, one big tournament where the two atheletes together participated. These examples demonstrate the value of our temporal Wikipedia indexes for temporal semantic research challenges.

\vspace{-0.2cm}
\section{Conclusions and Future Work}
\vspace{-0.2cm}
In this paper, we introduced Hedera, our ongoing work in supporting flexible and efficient access to Wikipedia revision history dataset. Hedera can work directly with raw data in the low-level, it uses Map-Reduce to achieve the high-performance computation. We open-source Hedera for future use in research communities, and believe our system is the first in public of this kind. Future work includes deeper integration with knowledge bases, with more API and services to access the extraction layer more flexibly.

\textbf{Acknowledgements}

\small{This work is partially funded by the FP7 project ForgetIT (under grant No. 600826) and the ERC Advanced Grant ALEXANDRIA (under grant No. 339233).}

\vspace{-0.2cm}

\bibliographystyle{abbrv}
\begin{tiny}

\end{tiny}

\end{document}